# Evolutionary Multi-Objective Optimization Algorithm Framework with Three Solution Sets

Hisao Ishibuchi*, Lie Meng Pang, and Ke Shang
Guangdong Provincial Key Laboratory of Brain-inspired Intelligent Computation
Department of Computer Science and Engineering, Southern University of Science and Technology
Shenzhen, China
hisao@sustech.edu.cn, panglm@sustech.edu.cn, kshang@foxmail.com

*Abstract*—It is assumed in the evolutionary multi-objective optimization (EMO) community that a final solution is selected by a decision maker from a non-dominated solution set obtained by an EMO algorithm. The number of solutions to be presented to the decision maker can be totally different. In some cases, the decision maker may want to examine only a few representative solutions from which a final solution is selected. In other cases, a large number of non-dominated solutions may be needed to visualize the Pareto front. In this paper, we suggest the use of a general EMO framework with three solution sets to handle various situations with respect to the required number of solutions. The three solution sets are the main population of an EMO algorithm, an external archive to store promising solutions, and a final solution set which is presented to the decision maker. The final solution set is selected from the archive. Thus the population size and the archive size can be arbitrarily specified as long as the archive size is not smaller than the required number of solutions. The final population is not necessarily to be a good solution set since it is not presented to the decision maker. Through computational experiments, we show the advantages of this framework over the standard final population and final archive frameworks. We also discuss how to select a final solution set and how to explain the reason for the selection, which is the first attempt towards an explainable EMO framework.

## I. Introduction

A variety of evolutionary multi-objective optimization (EMO) algorithms have been proposed in the last three decades [1]. When an EMO algorithm is designed, it is assumed that a final solution is selected by a decision maker from a set of obtained non-dominated solutions. For example, in Deb [1], multi-objective optimization is explained as a two-step procedure: The first step is the search for a number of non-dominated solutions by an EMO algorithm, and the second step is the choice of a final solution by the decision maker. The number of solutions to be presented to the decision maker can be totally different depending on the situation. In one situation where the decision maker needs to manually examine each solution for the final decision making, he/she does not want to examine many solutions [2]. This may be related to our biological limitation about the amount of information we can handle simultaneously, which is known as "the magical number seven plus or minus two" [3]. In another situation where the decision maker is trying to extract useful information about the shape of the Pareto front of a many-objective problem using visualization tools [4], a large number of non-dominated solutions may be needed. These discussions suggest that the number of solutions to be presented to the decision maker has a wide range from a very small number (e.g., 5-10 solutions) to a very large number (e.g., 500,000-1,000,000 solutions).

When an EMO algorithm has no archive, the final population is presented to the decision maker. In this final population framework, the required number of solutions to be presented to the decision maker is usually used as the population size. This framework has at least the following two difficulties:

(i) An appropriate specification of the population size is algorithm-dependent and problem-dependent [5]. In general, EMO algorithms do not always work well over a wide range of population size settings. If the population size is too small or too large, the performance of EMO algorithms can be severely deteriorated.

(ii) The final population is not always the best subset of the examined solutions during the execution of an EMO algorithm [5]. As reported in Li & Yao [6], it is often the case that many solutions in the final population are dominated by examined and deleted solutions in previous generations.

When an EMO algorithm has an archive, the final archive is usually presented to the decision maker. In this final archive framework, the archive size is the same as the required number of solutions. Since the population size can be arbitrarily specified, we can avoid the first difficulty. However, we still have the following difficulty (which is related to the second difficulty due to the limited size of the archive especially when the archive size is small):

* Corresponding Author: Hisao Ishibuchi
This work was supported by National Natural Science Foundation of China (Grant No. 61876075), Guangdong Provincial Key Laboratory (Grant No. 2020B121201001), the Program for Guangdong Introducing Innovative and Enterpreneurial Teams (Grant No. 2017ZT07X386), Shenzhen Science and Technology Program (Grant No. KQTD2016112514355531), the Program for University Key Laboratory of Guangdong Province (Grant No. 2017KSYS008).

(iii) The final archive is not always the best subset of the examined solutions. A better solution set can be obtained by selecting a subset of the examined solutions [7], [8].

Based on these discussions, we suggest the use of an EMO framework shown in Fig. 1 with three solution sets: the main population of an EMO algorithm, an archive, and a final solution set. The final solution set, which is selected from the archive after the execution of the EMO algorithm, is presented to the decision maker. In our former study [7], we proposed a special EMO algorithm framework with an unbounded archive. All the examined solutions are stored in the unbounded archive. The algorithm framework in [7] can be viewed as a special case of a more general framework in Fig. 1. When an EMO algorithm has an external archive which is stored separately from the main population (e.g., SPEA [9]), both of them are viewed as the main population (i.e., both of them are in the blue box in Fig. 1) in our algorithm framework since the external archive (i.e., the green box in Fig. 1) is used only for storing solutions externally from the EMO algorithm.

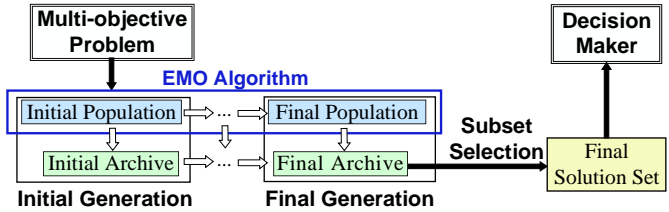

**Fig. 1.** The suggested EMO framework with three solution sets.

The suggested EMO framework in Fig. 1 has at least the following three advantages:

(a) Independent of the number of solutions to be presented to the decision maker, an appropriate population size can be arbitrarily specified in the EMO algorithm.

(b) The archive size can be also arbitrarily specified (as long as it is larger than or equal to the number of solutions to be presented to the decision maker), which includes the use of an unbounded archive.

(c) Since the final solution set is selected from the final archive, the final population does not have to be a good solution set. This increases the flexibility in the design of new EMO algorithms to efficiently search for the Pareto front.

The suggested framework is not a new idea. For example, in MOGLS [10] in 1990s, an unbounded archive was used to store all non-dominated solutions. An unbounded archive was also used for the performance evaluation [5], [6], [8], [11], [12] and the design of EMO algorithms [7]. The main contribution of this paper is to clearly demonstrate that the population size and the archive size should be specified independently of the required number of solutions. Our framework improves the applicability of existing EMO algorithms with respect to the required number of solutions. It also increases the flexibility in the design of new EMO algorithms.

One important issue in our framework is the selection of the final solution set. In the EMO community, solution subset selection based on the hypervolume indicator [13] has been actively studied under the name of hypervolume subset selection (HSS [14]). The HSS problem is to find a pre-specified number of solutions from a given solution set to maximize their hypervolume. Exact optimization algorithms [11], [15]-[18] as well as heuristic greedy algorithms [19] have been proposed. However, their use for decision making purposes has some difficulties. One is that the relation between the hypervolume maximization and the final solution selection is not clear. The hypervolume maximization does not necessarily mean the selection of promising solutions for decision making. We cannot clearly explain the reason why the hypervolume indicator should be used for selecting promising candidate solutions in decision making. Another difficulty is huge computation load when the final archive is large for a many-objective problem. It is also difficult to appropriately specify the reference point for hypervolume calculation [20].

In this paper, we propose the use of an expected loss function for subset selection to clearly explain the reason why the final solution set is selected for decision making. Let us denote the final archive by $S$. We need to select a subset $A$ with a pre-specified number (say, $k$) of solutions from $S$ (i.e., $A \subset S$ and $|A| = k$). A final solution is chosen from the subset $A$ by the decision maker. To select $A$, we formulate an expected loss function based on the following idea. Let $s$ be the (unknown) most preferred solution in $S$ by the decision maker. The loss of choosing $a$ in $A$ instead of $s$ in $S$ as the final solution is defined by the amount of the deterioration in the objective space caused by changing the choice from $s$ to $a$. It is assumed that $a$ is the solution with the minimum loss for $s$. If $s$ is included in $A$, the loss is zero since $s$ is selected (i.e., since $a = s$). The expected loss is formulated as the average loss over all solutions in $S$. The final solution set $A$ is selected from $S$ to minimize the expected loss. In this manner, we can clearly explain to the decision maker the reason why the final solution set is selected.

This paper is organized as follows. In Section II, we explain the suggested algorithm framework and its simple implementation with MOEA/D [21]. In Section III, we show the effectiveness of our framework through computational experiments on the DTLZ [22] and WFG [23] test problems with three and five objectives. In Section IV, we formulate the expected loss function. We also show that the formulated expected loss function is the same as the inverted generational distance plus (IGD+) indicator [24], which is a weakly Pareto compliant version of the IGD indicator [25]. In Section V, we discuss future research topics under our framework. Finally, we conclude this paper in Section VI.

II. PROPOSED FRAMEWORK AND ITS IMPLEMENTATION

An $m$-objective minimization problem is written as

$$\text{Minimize } f(x) = (f_1(x), f_2(x), ..., f_m(x)), \quad (1)$$
$$\text{subject to } x \in X, \quad (2)$$

where $f_i(x)$ is the $i$th objective to be minimized ($i = 1, 2, ..., m$), $x$ is a decision vector, and $X$ is the feasible region of $x$. Two solutions $x^A$ and $x^B$ are compared by the Pareto dominance relation as follows [1]: When $f_i(x^A) \leq f_i(x^B)$ for all $i$ and $f_j(x^A) < f_j(x^B)$ for at least one $j$, $x^A$ dominates $x^B$ (i.e., $x^A$ is better than

$x^B$). If $x^B$ is not dominated by any other solutions in $X$, $x^B$ is a Pareto optimal solution. The set of all Pareto optimal solutions is the Pareto optimal solution set. The projection of the Pareto optimal solution set to the objective space is the Pareto front. If all solutions in a solution set are non-dominated with each other, the solution set is a non-dominated solution set.

Various EMO algorithms have been proposed to search for a set of well-distributed non-dominated solutions. The obtained solution set is presented to the decision maker. Let us assume that we need to present $k$ solutions to the decision maker. Then, the task of an EMO algorithm is to search for a set of $k$ non-dominated solutions which well approximates the entire Pareto front of a given multi-objective problem.

In the standard final population framework, an EMO algorithm with the population size $k$ is applied to the given problem. Then, the final population is presented to the decision maker. Some EMO algorithms such as SPEA [9] and Two_Arch2 [26] have an archive. When such an EMO algorithm is used, the final archive is usually presented to the decision maker. Thus, the archive size is specified as $k$ in the final archive framework.

Recently, it was pointed out in some studies [5], [8], [11], [12] that the performance of EMO algorithms can be improved by selecting a subset of the examined solutions. It was also reported [6] that the final population often includes many solutions which are dominated by examined and deleted solutions in previous generations. Based on these studies, we suggested the use of an unbounded external archive in the design of new EMO algorithms in our former study [7]. Whereas this idea works well on many test problems (see [5], [11], [12]), it has at least three potential difficulties. One is that the archive size can become huge for a many-objective problem where almost all solutions are non-dominated. The second difficulty is that the update of the huge archive is time-consuming. The third difficulty is that the selection of the final solution set from the huge archive is not easy.

As we have already explained using Fig. 1, we suggest the use of an EMO framework with three solution sets: the main population, the archive, and the final solution set. When an EMO algorithm has its own archive, an additional external archive is used in our framework. Since the final solution set is selected from the final archive, the population size and the archive size can be arbitrarily specified independently of the required number of solutions. This is the main feature of our framework in comparison with the standard final population and final archive frameworks.

Here, we show a simple implementation example of our framework. We use MOEA/D [21] as an EMO algorithm and also for archive update. The final solution set is selected from the archive using heuristic distance-based greedy selection [19], [27]. The choice of these methods is for computational efficiency. Of course, we can use other EMO algorithms, archiving mechanisms and selection methods.

**Main EMO Algorithm**: In MOEA/D, a multi-objective problem is decomposed into a number of single-objective problems using a scalarizing function and a set of uniformly distributed weight vectors. For the $m$-objective problem in (1)-(2), all weight vectors $w = (w_1, w_2, ..., w_m)$ satisfying the following relations are generated:

$$w_1 + w_2 + ... + w_m = 1, \qquad (3)$$
$$w_i \in \{0/H, 1/H, 2/H, ..., H/H\}, \ i = 1, 2, ..., m, \qquad (4)$$

where $H$ is an integer to specify the total number of weight vectors (which is calculated as $_{m+H-1}C_{m-1}$ [21]). Since each weight vector has a single solution, the total number of weight vectors is the same as the population size in MOEA/D.

In MOEA/D, single-objective optimization of a scalarizing function is performed for each weight vector $w = (w_1, w_2, ..., w_m)$ in a collaborative manner. Three scalarizing functions (weighted sum, Tchebycheff and PBI) were examined in the original MOEA/D paper [21]. In this paper, we use the Tchebycheff and PBI functions (i.e., MOEA/D-TCH and MOEA/D-PBI). Since the weighted sum does not work well on non-convex test problems such as DTLZ and WFG, we do not use it in this paper. The Tchebycheff and PBI functions are defined in the same manner as in the original MOEA/D paper [21]. For example, the penalty parameter $\theta$ in the PBI function is specified as $\theta = 5$. The reference point in these functions is updated by the best value of each objective among all the examined solutions.

**Archive Update:** For archive update, we use a scalarizing function and weight vectors as in MOEA/D. The archive size is the same as the number of weight vectors. We use the same scalarizing function for archive update as in the main EMO part. After an initial population is generated in the main EMO part, the best solution with respect to the scalarizing function is assigned to each weight vector in the archive. After that, whenever a new solution is generated in the main EMO part, it is compared with all archive solutions. This comparison is based on the scalarizing function and the weight vector corresponding to each archive solution. When the new solution is better, the archive solution is updated with the new solution.

The weight vectors in the archive are generated by (3)-(4) as in MOEA/D. However, the archive size can be different from the population size. Even when the archive size is the same as the population size, the final archive is not always the same as the final population. This is because a new solution is compared with all solutions in the archive whereas it is compared with only the neighboring solutions in MOEA/D.

**Subset Selection:** Subset selection in our framework is to select a pre-specified number of solutions ($k$ solutions) from the final archive. Before subset selection, dominated solutions and duplicated solutions are removed from the final archive. For subset selection, we use heuristic distance-based greedy selection in Singh et al. [27]. First, an extreme solution with the best value of an objective is selected. In general, the final archive has $m$ extreme solutions of the given $m$-objective problem. One of them is randomly selected. The selected solution is added to the final solution set. Next, the most distant solution from the first solution in the objective space is selected and added as the second solution. Then, the distance from each of the remaining solutions in the final archive to the nearest solution in the final solution set is calculated, and the most distant solution is added to the final solution set as the third solution. In this manner, *k* solutions are selected. For comparison, we also use hypervolume-based greedy selection [19] where solutions are selected one by one to maximize the hypervolume of the selected solutions (i.e., to maximize the hypervolume of the final solution set). Of course, we can use other indicators and other subset selection methods as we discuss in Section IV.

### III. COMPUTATIONAL EXPERIMENTS

We apply MOEA/D-TCH and MOEA/D-PBI to DTLZ1-4 [22] and WFG1-9 [23] with three and five objectives under the following settings in the same manner as in other studies on evolutionary many-objective optimization (e.g., the NSGA-III paper [28] and the θ-DEA paper [29]):

Crossover: SBX with the probability 1.0 and the distribution index 30.
Mutation: Polynomial mutation with the probability 1/$D$ (where $D$ is the number of decision variables) and the distribution index 20.
Termination conditions:
  50,000 solution evaluations for the three-objective problems, and 200,000 solution evaluations for the five-objective problems.
Number of independent runs: 51 runs of each algorithm on each test problem.

For the population size and the archive size, we examine all the 16 combinations of their four settings in Table 1 (e.g., the population size 15 and the archive size 5985 for the five-objective problems). The "Standard" settings in Table 1 are frequently used in the literature (e.g., [28], [29]). The neighborhood size in MOEA/D is specified for each setting of the population size as follows: 15 (Small), 20 (Standard), 200 (Large), and 1000 (Very Large) for all test problems.

**Table 1.** Four settings of the population size and the archive size.

|  | Small | Standard | Large | Very Large |
|---|---|---|---|---|
| Three-objective | 15 | 91 | 990 | 5050 |
| Five-objective | 15 | 210 | 1001 | 5985 |

The performance of each algorithm for each combination of the population size and the archive size on each test problem is evaluated by the average hypervolume of the final archive over 51 runs. The final population is also evaluated by the average hypervolume. Hypervolume calculation is performed in the normalized objective space with the ideal point (0, ..., 0) and the nadir point (1, ..., 1). The reference point for hypervolume calculation is specified as (1.1, ..., 1.1), which is slightly worse than the nadir point (for the reference point specification, see [20]). Due to the page length limitation, we can show only a small part of our results. All results in the form of Fig. 2 (a) are available from https://github.com/HisaoLabSUSTC/EMO-Framework.

Experimental results by MOEA/D-TCH on the three-objective DTLZ1 problem are summarized in Fig. 2 (a) where the average hypervolume values of each archive and the main population are shown for the corresponding main population size (i.e., the horizontal axis). For example, the left-most blue square is the average result of the size 15 main population. The right-most yellow circle is the average result of the size 91 archive when it is used in MOEA/D-TCH with the size 5050 main population. In Fig. 2 (a), the performance of the main population is improved by increasing the population size. This does not mean that the search ability of MOEA/D-TCH is improved by increasing the population size. In Fig. 2 (a), each connected line is flat except for the blue square results of the main population. This means that the search ability of MOEA/D in Fig. 2 (a) is not sensitive to the population size. This observation cannot be obtained from the performance analysis of only the main population. Of course, a larger archive has a larger average hypervolume value in Fig. 2 (a).

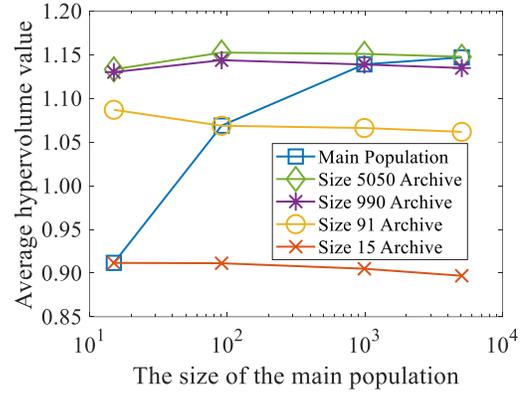

(a) Average hypervolume value.

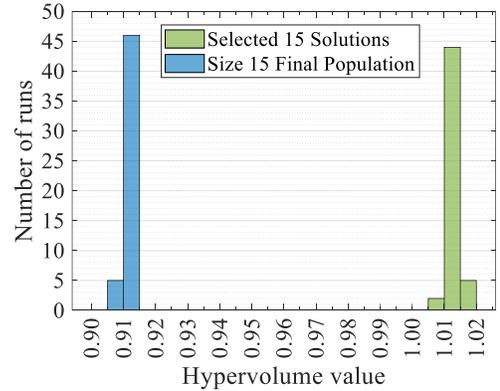

(b) Performance of the selected 15 solutions.

**Fig. 2.** Experimental results by MOEA/D-TCH on the three-objective DTLZ1 problem.

One may think that our framework does not improve the performance of MOEA/D-TCH in Fig. 2 (a) since the final population performance is the same as the archive performance with the same size (e.g., see the left-most blue square and the four red crosses). Actually, our framework can significantly improve the performance of MOEA/D-TCH by choosing the required number of solutions from the archive. For example, by selecting 15 solutions using the distance-based method from the size 5050 final archive from MOEA/D-TCH with the population size 15 (i.e., by choosing 15 solutions from the left-most green diamond result in Fig. 2 (a)), we obtain the average hypervolume value 1.0073. When we use the hypervolume-based selection method, we obtain a slightly better average hypervolume value 1.0133. These results are clearly better than the final population of the size 15 and the four final archives of the size 15 in Fig. 2 (a). The distribution of the hypervolume values of the selected 15 solutions by the hypervolume-based method in 51 runs is compared with that of the final population of the size 15 in Fig. 2 (b). Clearly larger hypervolume values are obtained by the selected 15 solutions in Fig. 2 (b). That is, clearly better results are obtained by our framework. The statistical significance test based on the Wilcoxon rank sum test confirms that the selected 15 solutions are better than the final population of size 15 with the 5% significance level.

With respect to the effect of the population size on the search ability of MOEA/D, we observe three typical patterns: (i) no clear relation as in Fig. 2 (a), (ii) positive correlation, and (iii) negative correlation. An example of positive correlation results is shown in Fig. 3 (a) where the best results are always obtained when the population size is 5050 (independent of the archive size).

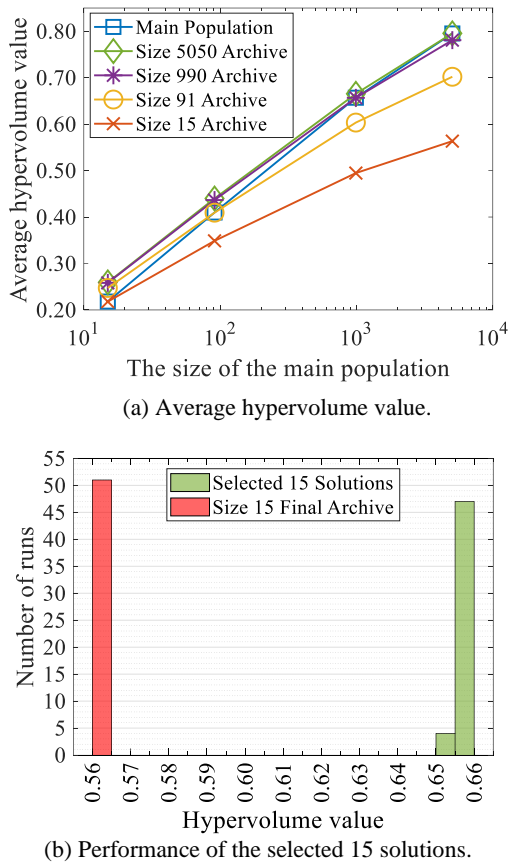

(a) Average hypervolume value.

(b) Performance of the selected 15 solutions.

**Fig. 3.** Experimental results by MOEA/D-TCH on the three-objective DTLZ4 problem.

Even when we need only 15 solutions, the best result is obtained from the population size 5050 in Fig. 3 (a). For example, the average hypervolume value of the final archive of the size 15 is 0.5634 in the case of the main population size 5050 (the right-most red cross in Fig. 3 (a)) whereas it is 0.2179 in the case of the main population size 15 (the left-most red cross). This observation shows the importance of using an appropriate population size independently of the required number of solutions. The final archive result 0.5634 (the right-most red cross) can be further improved by selecting 15 solutions from the final archive of the size 5050 in the case of the main population size 5050 (i.e., by the selection from the right-most green diamond result) to 0.6234 (by distance-based selection) and 0.6588 (by hypervolume-based selection). Fig. 3 (b) shows the distribution of the hypervolume values of the selected 15 solutions by the hypervolume-based method in 51 runs in comparison with that of the final archive of size 15 (the right-most red cross result). Even when the required number of solutions is very small (e.g., 15), Fig. 3 (b) clearly shows the usefulness of using a large archive together with a solution subset selection method.

Two examples of negative correlation are shown in Fig. 4 where the best population size is 15 even when we need more than 5000 solutions. That is, in each figure in Fig. 4, the best result is the left-most green diamond. If we examine only the blue square results (i.e., if we examine only the performance of the main population) in each figure in Fig. 4, we may have the following incorrect observations: Good results are not obtained from a too small or too large population, and the best results are obtained from the standard settings of the population size (i.e., 91 for the three-objective DTLZ3 problem and 210 for the five-objective WFG3 problem).

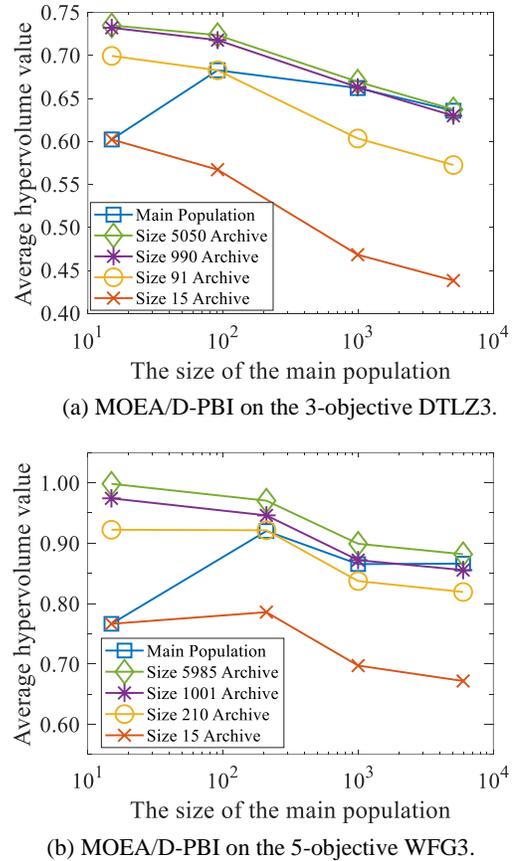

(a) MOEA/D-PBI on the 3-objective DTLZ3.

(b) MOEA/D-PBI on the 5-objective WFG3.

**Fig. 4.** Examples of negative correlation between the search ability and the population size.

One interesting observation in each figure in Fig. 4 is that the four connected lines for the archives (except for the blue square results of the main population) show a similar behavior whereas the archive size is totally different. This observation suggests that the best main population size does not depend on the archive size. The same observation is obtained from Fig. 2 (a), Fig. 3 (a) and almost all results on the other test problems (https://github.com/HisaoLabSUSTC/EMO-Framework). To the best of our knowledge, Figs. 2-4 are the first clear examinations about the relation between the search ability of EMO algorithms and the population size using external archives of various sizes. As we explained for Fig. 2 (a) and Fig. 4, the performance evaluation of only the final population leads to incorrect observations about this relation.

## IV. SOLUTION SUBSET SELECTION

One important component in our framework is a selection method of the final solution set. A related important issue from a decision making viewpoint is how to explain the reason for the selection of the final solution set. In this section, we discuss the second issue, which has not been discussed in many studies. Let us consider the following case. For a given multi-objective problem, a large number of non-dominated solutions (e.g., 500 solutions) have already been obtained by an EMO algorithm. A decision maker wants to examine only a small number of promising solutions (e.g., 10 solutions) to choose a final solution. He/she also wants to receive a clear and convincing explanation about the reason for the selection of the 10 promising solutions from the 500 original candidate solutions.

As the first attempt towards the design of an explainable EMO-based decision making framework, we formulate an expected loss function in order to give a clear meaning to solution subset selection from a decision making viewpoint. We assume that we have a set of $n$ non-dominated solutions in the $m$-dimensional objective space: $S = \{s_1, s_2, ..., s_n\}$. Our problem is to select a subset $A$ of size $k$ from $S$ (i.e., $A = \{a_1, a_2, ..., a_k\} \subset S$).

Let $s$ be the (unknown) most preferred solution by the decision maker in the solution set $S$. If $s$ is in the selected solution subset $A$, it will be chosen as the final solution by the decision maker. In this case, we have no loss by the selection of $A$ from $S$. Since the number of selected solutions in $A$ is usually much smaller than the total number of solutions in $S$ (i.e., $k = |A| << |S|$), it is likely that $s$ is not included in $A$. In this case, we assume that the nearest solution $a$ in $A$ to $s$ will be chosen as the final solution by the decision maker. To formulate the loss by the selection of $A$ in this context, we need to discuss the following two issues: the choice of the nearest solution $a$ and the definition of the loss by choosing $a$ (in the selected candidate solution set $A$) instead of $s$ (in the original candidate solution set $S$).

First we discuss the second issue. In Fig. 5, we show three cases of the locations of $s = (s_1, s_2)$ and $a = (a_1, a_2)$ in the objective space of a two-objective minimization problem. In Fig. 5 (a), $a$ is dominated by $s$. In this case, we use the distance between $a$ and $s$ as the loss by choosing $a$ instead of $s$ (actually this case does not happen since all solutions in $S$ are non-dominated). In Fig. 5 (b), $a$ and $s$ are non-dominated. Since $a$ is better than $s$ for $f_2(x)$, their distance with respect to $f_2(x)$ is not the loss. The loss is their distance with respect to $f_1(x)$ in Fig. 5 (b). On the contrary, in Fig. 5 (c), $a$ is better than $s$ with respect to $f_1(x)$. Thus, the loss is their distance only for $f_2(x)$.

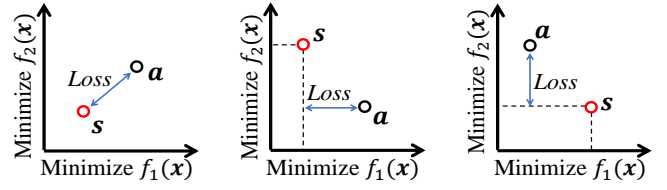

(a) $s_1 < a_1$ and $s_2 < a_2$.  (b) $s_1 < a_1$ and $s_2 > a_2$.  (c) $s_1 > a_1$ and $s_2 < a_2$.

**Fig. 5.** Three cases of the locations of $s$ and $a$ for a two-objective minimization problem.

From the above discussions, we define the loss by choosing $a = (a_1, a_2)$ instead of $s = (s_1, s_2)$ for the two-objective minimization problem as follows:

$$Loss(\boldsymbol{a}, \boldsymbol{s}) = \sqrt{(\max\{0, a_1 - s_1\})^2 + (\max\{0, a_2 - s_2\})^2} \ . \quad (5)$$

This formulation is generalized to the case of $m$ objectives as follows:

$$Loss(\boldsymbol{a}, \boldsymbol{s}) = \sqrt{\sum_{i=1}^{m}(\max\{0, a_i - s_i\})^2} \ . \quad (6)$$

In this formulation, the distance for each objective with $a_i < s_i$ is ignored since $a$ is better than $s$ with respect to such an objective. For maximization problems, $a_i - s_i$ is replaced with $s_i - a_i$ in (6).

Next we discuss the choice of a single final solution from the subset $A$ by the decision maker whose preferred solution among $S$ is $s$. It may be reasonable to assume that the decision maker will choose the solution with the minimum loss from $A$. Based on this assumption, we define the loss by selecting the subset $A$ from $S$ as follows:

$$Loss(A, \boldsymbol{s}) = \min_{\boldsymbol{a} \in A}\{Loss(\boldsymbol{a}, \boldsymbol{s})\} . \quad (7)$$

This formulation defines the loss by the selection of the solution subset $A$ for the decision maker whose preferred solution is $s$. However, we have no information about the preferred solution $s$. Thus we assume that each solution $s$ in $S$ has the same probability to be selected as the final solution. Based on this assumption, we define the expected loss by the selection of the solution subset $A$ from $S$ as

$$Loss(A, S) = \frac{1}{|S|}\sum_{s \in S}\min_{\boldsymbol{a} \in A}\{Loss(\boldsymbol{a}, \boldsymbol{s})\} . \quad (8)$$

Our proposal is to select the solution subset that minimizes the expected loss function in (8). Using (6), (8) is rewritten as

$$Loss(A, S) = \frac{1}{|S|}\sum_{s \in S}\min_{\boldsymbol{a} \in A}\left\{\sqrt{\sum_{i=1}^{m}(\max\{0, a_i - s_i\})^2}\right\} \quad (9)$$

This is the same as the IGD$^+$ indicator [24] of the solution subset $A$ when the given solution set $S$ is used as the reference point set:

$$Loss(A, S) = IGD^+(A, S) = \frac{1}{|S|} \sum_{s \in S} \min_{a \in A} \left\{ \sqrt{\sum_{i=1}^{m} (\max\{0, a_i - s_i\})^2} \right\}. \quad (10)$$

The IGD$^+$ indicator is a modified version of the IGD indicator [25]. One advantage of IGD$^+$ over IGD is that IGD$^+$ is weakly Pareto compliant whereas IGD is not Pareto compliant (for the Pareto compliance property, see Zitzler et al. [30]).

From these discussions, we can see that the IGD$^+$-based subset selection has a clear explanation: The selected subset has the minimum expected loss when the given solution set $S$ (i.e., the final archive in our EMO framework) is used as the reference point set. In general, the choice of an appropriate reference point set is difficult in the application of the GD, IGD and IGD$^+$ indicators [31]. In our EMO framework, we can obtain a large number of non-dominated solutions as reference points by using a large archive. This is another advantage of our EMO framework.

## V. Future Research Topics

Our EMO framework poses a number of interesting future research topics as follows:

**Performance Evaluation of Existing EMO Algorithms**: As shown by our experimental results, the performance of the final population can be significantly improved by selecting the required number of solutions from a large archive. It was also shown that the performance of EMO algorithms strongly depends on the specification of the population size. However, EMO algorithms have been compared using the performance of the final population with a pre-specified common population size in the literature. It is likely that those comparison results do not show the true performance of each EMO algorithm. Performance re-evaluation of existing EMO algorithms under our framework will lead to different comparison results.

**Proposal of New EMO Algorithms**: Since the final population does not have to be a good solution set in our framework, we have a large flexibility in the design of new EMO algorithms. For example, in a population size adaptation mechanism, the final population size can be very small. In a weight vector adjustment mechanism, the final weight vectors can be heavily biased towards some difficult regions of the Pareto front. Recently, it has been clearly demonstrated that the best MOEA/D algorithm configurations are different between the final population framework and the solution selection framework with an unbounded external archive [32].

**Efficient Archive Update**: Since we can arbitrarily specify the archive size independently of the required number of solutions (including the use of an unbounded archive), an efficient archive update mechanism may be needed for a huge archive. It is also an interesting research topic to compare the efficiency among the following three strategies of archive update for choosing non-dominated solutions from all the examined solutions: every generation update, periodical update (e.g., every 50 generations) and update only after the execution of an EMO algorithm [33].

**Explainable Subset Selection**: Whereas we proposed the use of the expected loss as an explainable subset selection criterion, further discussions are needed about the validity of its use for decision making purposes. For example, in its formulation, the gain by changing the choice from *s* to *a* is simply ignored. By combining the expected gain with the expected loss, we may be able to formulate a more acceptable criterion for the decision maker.

## VI. Conclusions

In this paper, first we proposed a general EMO framework with three solution sets: the main population of an EMO algorithm, the archive and the final solution set. Next, we clearly demonstrated that the population size and the archive size should be specified independently of the required number of solutions through computational experiments. Our experimental results showed that the search ability of EMO algorithms cannot be evaluated by the performance of the final population. Then, we formulated the expected loss function as an explainable criterion for the final solution selection. Finally, we suggested some future research topics. We believe that the proposed framework will improve the performance and the applicability of existing EMO algorithms. It will also work as a platform for the design of new EMO algorithms.